\pdfoutput=1

\documentclass[11pt]{article}

\usepackage[final]{acl}

\usepackage{times}
\usepackage{latexsym}
\usepackage{fancyvrb}
\usepackage{natbib}
\usepackage[T1]{fontenc}

\usepackage[utf8]{inputenc}

\usepackage{microtype}

\usepackage{inconsolata}

\usepackage{graphicx}
\usepackage{tabularx}
\usepackage{multirow}
\usepackage{booktabs,amsfonts,dcolumn}
\usepackage[export]{adjustbox}
\usepackage{caption}

\usepackage{xcolor}
\usepackage[newfloat,frozencache,cachedir=.]{minted}
\usepackage{enumitem}
\SetupFloatingEnvironment{listing}{placement=t!}    

\usepackage{cleveref}

\newcommand{\ragf}{\textsc{RAG~Foundry}}

\title{RAG~Foundry: A Framework for Enhancing LLMs for Retrieval Augmented Generation}

\author{Daniel Fleischer \qquad Moshe Berchansky \qquad Moshe Wasserblat \qquad Peter Izsak
\vspace{0.3em} \\
Intel Labs \\ 
\tt \normalsize \{daniel.fleischer, moshe.berchansky, moshe.wasserblat, peter.izsak\}@intel.com
}

\begin{document}

\maketitle

\begin{abstract}

Implementing Retrieval-Augmented Generation (RAG) systems is inherently complex, requiring deep understanding of data, use cases, and intricate design decisions.  
Additionally, evaluating these systems presents significant challenges, necessitating assessment of both retrieval accuracy and generative quality through a multi-faceted approach.
We introduce \ragf{}, an open-source framework for augmenting large language models  for RAG use cases.
\ragf{} integrates data creation, training, inference and evaluation into a single workflow, facilitating the creation of data-augmented datasets for training and evaluating large language models in RAG settings.
This integration enables rapid prototyping and experimentation with various RAG techniques, allowing users to easily generate datasets and train RAG models using internal or specialized knowledge sources.
We demonstrate the framework effectiveness by augmenting and fine-tuning Llama-3 and Phi-3 models with diverse RAG configurations, showcasing consistent improvements across three knowledge-intensive datasets.
Code is released as open-source in \mbox{\small\url{https://github.com/IntelLabs/RAGFoundry}}.









\end{abstract}

\begin{figure}
    \centering
    \includegraphics[width=1\linewidth]{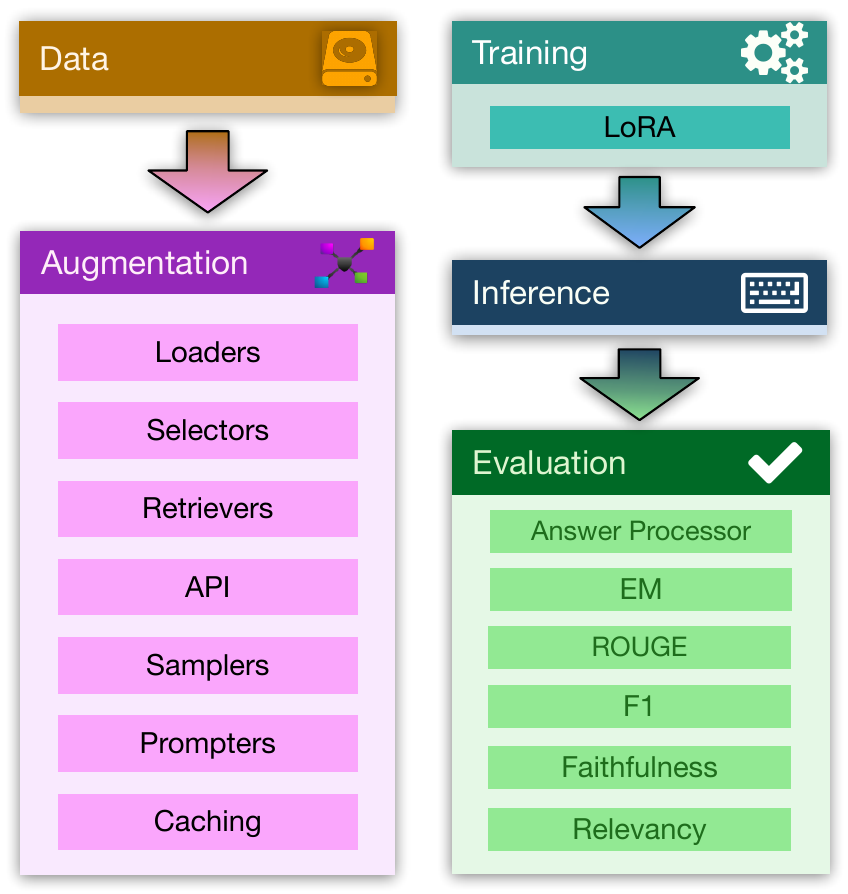}
    \caption{An overview of the \ragf{} framework: the \textbf{Data Augmentation} module persists RAG interactions into a dedicated dataset, which is then used for training, inference and evaluation.}
    \label{fig:enter-label}
\end{figure}

\section{Introduction}

Large Language Models (LLMs) have emerged as a transformative force in the field of AI, demonstrating an impressive ability to perform a wide range of tasks that traditionally required human intelligence \cite{brownLanguageModelsAre2020,kojimaLargeLanguageModels2022}.
Despite their impressive capabilities, LLMs have inherent limitations. These models can produce plausible-sounding but incorrect or nonsensical answers, struggle with factual accuracy, lack access to up-to-date information after their training cutoff and struggle in attending to relevant information in large contexts \cite{huangSurveyHallucinationLarge2023, liu2023lostmiddlelanguagemodels}.
Retrieval-Augmented Generation (RAG) enhances LLMs performance by integrating external information using retrieval mechanisms. Combining retrieval that leverages vast knowledge-bases outside the \textit{knowledge} of the model, effectively addresses knowledge limitations, can reduce hallucinations, improve the relevance of generated content, provide interpretability and could be vastly more cost-efficient \cite{lewisRetrievalAugmentedGenerationKnowledgeIntensive2021,Mallen2022WhenNT,gaoRetrievalAugmentedGenerationLarge2023,asai2023selfraglearningretrievegenerate,Borgeaud2021ImprovingLM,pengCheckYourFacts2023b,dejongPrecomputedMemoryOnthefly2023a}. Furthermore, recent research indicates that fine-tuning LLMs for RAG can achieve state-of-the-art performance, surpassing that of larger, proprietary models \cite{yuRankRAGUnifyingContext2024,liuChatQASurpassingGPT42024}.

However, the implementation of RAG systems is inherently complex and requires a series of intricate decisions that can significantly impact the performance of the system. This process demands a thorough understanding of the data and use case, and often, solutions do not generalize well to other domains \cite{barnett2024sevenfailurepointsengineering,balaguerRAGVsFinetuning2024}.
Some key RAG design decisions include text embedding, indexing parameters, retrieval algorithms, query building, and prompt design, among other considerations beyond the LLM configuration \cite{wangSearchingBestPractices2024}.
Another issue is reproducibility: achieving consistent and comparable results across runs, datasets and tasks. Variations in training data, pre-processing steps, model configurations, and hardware can lead to discrepancies in performance, making it challenging for researchers and practitioners to replicate findings and build upon previous work.
Additionally, evaluating RAG systems presents a challenge due to the dual reliance on retrieval accuracy and generative quality. These systems require a sophisticated evaluation suite that accounts for the interplay among the retrieved information, the formalization of data, and the generated output \cite{chenBenchmarkingLargeLanguage2023a,yuEvaluationRetrievalAugmentedGeneration2024,es-etal-2024-ragas}.

We introduce \ragf, an open-source python framework for developing sophisticated retrieval-augmented LLMs for RAG use-cases.
The library supports researchers and practitioners in the nuanced task of enhancing the capabilities of LLMs in RAG use cases. It is highly customizable, facilitating rapid prototyping and experimentation across all aspects of RAG, including data selection, aggregation and filtering, retrieval, text processing, document ranking, few-shot generation, prompt design using templates, fine-tuning, inference, and evaluation. To cater to the specific needs of researchers, we designed the framework to function as an end-to-end experimentation environment. The backbone of the library consists of four distinct modules: data creation, training, inference, and evaluation. Each module is encapsulated and controlled by a configuration file, ensuring compatibility between the output of one module and the input of the next. This modular approach allows each step to be isolated and independently experimented with, enabling the production of multiple outputs and the concurrent execution of numerous experiments. Evaluation can be conducted on the generated outputs as well as on any feature within the data, including retrieval, ranking, and reasoning.

To illustrate the utility of the framework, we conducted experiments involving retrieval, fine-tuning, chain-of-thought (CoT) reasoning \cite{wu-etal-2023-chain} and a negative distractor-documents technique \cite{raft2024}. We compared two widely accepted baseline models using various enhancement methods across three knowledge-intensive question-answering tasks, demonstrating the effectiveness of \ragf.

\section{Related Work}

There are numerous open-source tools related to the different aspects of RAG, namely inference, training and evaluation. LlamaIndex \cite{Liu_LlamaIndex_2022}, LangChain \cite{Chase_LangChain_2022} and Haystack \cite{Pietsch_Haystack_the_end-to-end_2019} are well known libraries for composing RAG pipelines; however they are not focused on evaluation and their training capability is under-developed. 

\citet{hoshiRaLLeFrameworkDeveloping2023} proposes a framework for developing RAG-based LLMs; while our processing may be similar in the sense of being comprised of custom individual steps, they do not introduce any form of training. \citet{khattab2023dspy,khattab2022demonstrate} presents a different approach, where LLM prompting is represented as a programming language, to be optimized and compiled; a rather unique and general approach that could benefit RAG but has a high level of complexity due to the abstractions introduced. \citet{saad-falconARESAutomatedEvaluation2024} focuses more on the evaluation aspect, by creating synthetic data and training an LLM critic to evaluate the RAG system. \citet{hsiaRAGGEDInformedDesign2024} studies aspects of retrieval on the performance of RAG; our RAG Foundry library is general and enables experimentation on all aspects of RAG: retrieval, text-processing, prompt design, model selection, inference and evaluations. 

Recently, a concurrent work by \citet{jinFlashRAGModularToolkit2024} proposes a RAG building framework, including some RAG implementations and datasets; we focus on extensibility, letting users define custom types of pipelines with custom components. \citet{rauBERGENBenchmarkingLibrary2024} presents a framework, sharing a similar design-principle of extensibility-through-configuration as ours; their library imposes a specific workflow structure (retriever, ranker, LLM) while our library is more general and does not imposes any specific paradigm.


\section{RAG~Foundry}

\begin{listing}
    \begin{minted}[fontsize=\footnotesize,frame=lines,breaklines]{yaml}
name: my_pipeline
cache: true
steps:
  - _target_: dataset_loaders.loaders.HFLoader
    inputs: main
    dataset_config:
      path: "Tevatron/wikipedia-trivia"
      split: train

  - _target_: dataset_loaders.loaders.LocalLoader
    inputs: fewshot-data
    filename: prepared-fewshot-data.jsonl

  - _target_: global_steps.sampling.ShuffleSelect
    inputs: main
    shuffle: 42
    limit: 10000

  - _target_: local_steps.retrievers.HaystackRetriever
    inputs: main
    pipeline_path: configs/qdrant.yaml
    query_key: query
    docs_key: positive_passages

  - _target_: global_steps.sampling.FewShot
    inputs: main
    input_dataset: fewshot-data
    k: 3
    output_key: fewshot_examples

  - _target_: local_steps.prompter.TextPrompter
    inputs: main
    prompt_file: prompts/basic.txt
    output_key: my_prompt
    mapping:
      question: query
      context: positive_passages
      fewshot: fewshot_examples
      answer: answers

  - _target_: global_steps.output.OutputData
    inputs: main
    file_name: TQA_train_processed.jsonl
    \end{minted}
    \caption{Example of a dataset creation configuration. The example contains data loading, shuffling, sampling, retrieval, few-shot collection, prompt building and saving steps.}
    \label{fig:main_processing}
\end{listing}

The \ragf{} framework facilitates rapid prototyping and experimentation with various RAG settings and configurations. The library is composed of four modules: dataset creation, training, inference, and evaluation. Below, we expand on each of the modules and provide example configurations for running them.


\subsection{Data Creation and Processing}

The \textit{processing} module facilitates the creation of context-enhanced datasets by persisting RAG interactions, which are essential for RAG-oriented training and inference \cite{berchansky2024cotarchainofthoughtattributionreasoning,liuChatQASurpassingGPT42024,yuRankRAGUnifyingContext2024}. These interactions encompass dataset loading, column normalization, data aggregation, information retrieval, template-based prompt creation, and various other forms of pre-processing. The processed data can be saved in a consistent, model-independent format, along with all associated metadata, ensuring compatibility and reproducibility across different models and experiments.

The processing module is comprised of an abstract pipeline with multiple steps, each defined by Python classes that implement specific data processing functionalities. These steps are categorized into two types:

\begin{itemize}[noitemsep,topsep=4pt,parsep=0.4em,leftmargin=*]
    \item \textbf{Global Steps}:  Can act on the dataset as a whole, making them useful for operations such as aggregations, group-by, examples filtering, join operations, and more. 
    \item \textbf{Local Steps}: Operate on individual examples, making them suitable for tasks such as retrieval, text processing, and field manipulation.
\end{itemize}

The modular design allows for building flexible and efficient data processes, tailored to the needs of RAG-oriented training and inference. Steps can be categorized into the following non-exclusive categories:

\begin{itemize}[noitemsep,topsep=4pt,parsep=0.4em,leftmargin=*]
\item \textbf{Loaders}: Load datasets from the Hugging Face\footnote{\url{https://huggingface.co/}} hub or from local sources.

\item \textbf{Selectors}: Filter examples, shuffle datasets, and select subset datasets.

\item \textbf{Retrievers}: Integrate information from external databases, tools, libraries and pipelines.

\item \textbf{Samplers}: Collect random examples or features from any dataset to compile few-shot or negative examples.

\item \textbf{Prompters}: Format prompts using custom templates and keyword mappings.

\end{itemize}

The processing module supports the handling of \textbf{multiple datasets} at once, through global dataset sharing. This feature allows each step of the pipeline to access any of the loaded datasets, enhancing flexibility and allowing for complex processing procedures. Furthermore, the module includes \textbf{step caching}, which caches each pipeline step locally. This improves compute efficiency, and facilitates easy reproduction of results.






\subsubsection{Example: Enhancing a Q\&A Dataset}
To showcase the effectiveness of the processing module, we demonstrate how to enrich a question-answering dataset with external information fetched using a retrieval pipeline, prepare few-shot examples and combine everything together using a prompt template. 
\Cref{fig:main_processing} demonstrates how such a processing pipeline is defined using a YAML configuration. 
The main structure of the file is a list of steps, each defined by a \verb|_target_| which points to the step implementation. Each step has \verb|inputs|, which is a name or list of dataset names to act upon. Other keys in a step  relate to specific step logic. 

The first two steps in \cref{fig:main_processing} load datasets from Hugging Face hub and from a local path. The third step shuffles and selects 10k examples from the main dataset. The forth step runs a Haystack-based \cite{Pietsch_Haystack_the_end-to-end_2019} retrieval pipeline to retrieve relevant passages using questions from the loaded dataset as queries, storing them in \verb|docs_key|. We note that different retrieval processes or frameworks \cite{Liu_LlamaIndex_2022, Chase_LangChain_2022, Lin_etal_SIGIR2021_Pyserini} can be used in retrieval steps. 
The fifth step selects 3 few-shot examples from the secondary dataset, following a prompt generator step that loads a prompt template and replaces all given information according to the defined \verb|mapping| dictionary. Lastly, the dataset is saved to a local path.




\begin{listing}
\begin{minted}[fontsize=\footnotesize,frame=lines,breaklines]{yaml}
model:
  _target_: ragfoundry.models.hf.HFTrain
  model_name_or_path: "microsoft/Phi-3-mini-128k-instruct"
  load_in_8bit: true
  lora:
    peft_type: "LORA"
    r: 16
    target_modules: ["qkv_proj"]
  completion_start: "<|assistant|>"

train:
  gradient_accumulation_steps: 4
  learning_rate: 2e-05
  lr_scheduler_type: "cosine"
  num_train_epochs: 1
  optim: "paged_adamw_8bit"

instruction: prompts/prompt_instructions/qa.txt
data_file: TQA_train_processed.jsonl
\end{minted}
\caption{Example of a training configuration. Model and training parameters are specified, in addition to an instruction file containing the system prompt.}
    \label{fig:training}
\end{listing}

\subsection{Training}

We provide a \textit{training} module to fine-tune models given the datasets created by the previous processing module. The training module relies on the well established training framework TRL\footnote{\url{https://github.com/huggingface/trl}} and supports advanced and efficient training techniques, e.g. LoRA~\cite{huLoRALowRankAdaptation2021}. An example of a training configuration is presented in \cref{fig:training}.




\subsection{Inference}

The \textit{inference} module generates predictions given the processed datasets created by the processing module. Inference is conceptually separated from the evaluation step, since it is more computationally demanding than evaluation.
Additionally, one can run multiple evaluations on a single, prepared inference results file.
An example configuration for generating predictions given a dataset is presented in \cref{fig:inference}. 


\begin{listing}
\begin{minted}[fontsize=\footnotesize,frame=lines,breaklines]{yaml}
model:
  _target_: ragfoundry.models.hf.HFInference
  model_name_or_path: "microsoft/Phi-3-mini-128k-instruct"
  load_in_8bit: true
  instruction: prompts/prompt_instructions/qa.txt
  lora_path: /path/to/adapter
  generation:
    do_sample: false
    max_new_tokens: 50
    return_full_text: false

data_file: my-processed-data.jsnol
generated_file: model-predictions.jsonl
\end{minted}
\caption{Example of an inference configuration. In addition to model and generation options, a system prompt can be defined.}
    \label{fig:inference}
\end{listing}

\subsection{Evaluation}

The goal of the framework is augmenting LLMs for RAG. The \textit{evaluation} module allows users to run collections of metrics to evaluate RAG techniques and tuning processes.
The evaluation module loads the output of the inference module and runs a configurable list of metrics.
Metrics are classes implemented in the library. These classes can be as simple as wrappers around other evaluation libraries, or can be implemented by the user. \textbf{Local metrics} can be run on individual examples, like Exact Match (EM), while \textbf{Global metrics} run on the entire dataset as a whole, e.g. Recall (for classification-based metrics). Metrics can use any field and metadata in the dataset, not just the input-output pairs. Some of the metrics implemented in the library include: a wrapper for the Hugging Face \textit{evaluate} library, EM,  F1, classification metrics, BERTScore
\cite{zhangBERTScoreEvaluatingText2019}, Semantic Similarity and a wrapper for DeepEval\footnote{\url{https://github.com/confident-ai/deepeval}} (for using the RAGAS metrics \cite{es-etal-2024-ragas}). After the evaluation is completed, a results file is written to disk with the local and global metrics results.



Furthermore, the evaluation module uses a processing step called an \textbf{Answer Processor}, which can implement custom logic and serve many purposes, including cleaning and aligning outputs; for example, using regex, one can isolate answers, remove stop words, chain-of-thought reasoning, define a stopping criteria, process citations and attributions and any other form of processing needed for a given evaluation.

See \cref{fig:evaluation} for a configuration example; it contains an answer processor that extracts an answer from an output, and a list of metrics to run.


\begin{listing}
\begin{minted}[fontsize=\footnotesize,frame=lines,breaklines]{yaml}
answer_processor:
  _target_: ragfoundry.processing.RegexAnswer
  capture_pattern: "Answer: (.*)"
  stopping_pattern:

metrics:
  - _target_: ragfoundry.evaluation.HFEvaluate
    metric_names: ["rouge"]
  - _target_: ragfoundry.evaluation.EM
  - _target_: ragfoundry.evaluation.F1
  - _target_: ragfoundry.evaluation.BERTScore
    model: "microsoft/deberta-large-mnli"
  - _target_: ragfoundry.evaluation.Faithfulness
  - _target_: ragfoundry.evaluation.Relevancy
    embeddings: "BAAI/bge-small-en-v1.5"

results_file: my-evaluation.yaml
generated_file: model-prediction.jsonl
data_file: my-processed-data.jsonl
\end{minted}
\caption{Example of an evaluation configuration; it contains an answer processor, as well as the list of metrics, with optional parameters, to run.}
    \label{fig:evaluation}
\end{listing}

\section{Experiments: RAG Tuning}

To illustrate the usage and usefulness of the \ragf{} library, we experiment with several possible RAG improvements to LLMs, and evaluate the results on three knowledge-intensive tasks. 


\subsection{RAG Augmentation Techniques}

We explore several techniques for RAG augmentation, and use \ragf{} to easily implement and evaluate their benefit. As an initial step, we evaluate unmodified models; we set \textbf{Baseline} as a configuration that is defined by running unmodified models and without any external knowledge. We define a \textbf{RAG} setting that introduces top-relevant documents in a consistent prompt template format with a system instruction, and a \textbf{CoT} scheme which guides the model to use the retrieved context, explain the steps, quote relevant parts and produce a final answer.
Complementing that, we explore fine-tuning recipes. We fine-tune the model in the \textbf{RAG} setup and denote is as \textbf{RAG-sft}. To complement \textbf{CoT}, we implemented a fine-tuning recipe, denoted as \textbf{CoT-sft}, introduced in \cite{raft2024}, where gold documents and purely distractor documents are used in the prompt, determined by probability, in conjunction with a CoT prompt. All prompt templates are included in \cref{appendix:prompts}.


\subsection{Datasets}
We evaluate our models on \mbox{\textbf{TriviaQA}} \cite{joshiTriviaQALargeScale2017}, \textbf{PubmedQA} \cite{jinPubMedQADatasetBiomedical2019}, and \textbf{ASQA} \cite{stelmakhASQAFactoidQuestions2022} which are knowledge intensive question-answering datasets which benefit from external sources. The TriviaQA and PubmedQA datasets contain relevant context; for ASQA, retrieval was done over a Wikipedia corpus using a dense retriever\footnote{\href{https://huggingface.co/BAAI/llm-embedder}{BAAI/llm-embedder}}. Dataset sources and sizes are included in \cref{appendix:datasets}.

\subsection{Models}
We experiment with two representative models: Llama-3\footnote{\href{https://huggingface.co/meta-llama/Meta-Llama-3-8B-Instruct}{meta-llama/Meta-Llama-3-8B-Instruct}.} \cite{llama1,llama3modelcard} and Phi-3\footnote{\href{https://huggingface.co/microsoft/Phi-3-mini-128k-instruct}{microsoft/Phi-3-mini-128k-instruct}.} \cite{phi3technicalreporthighly} as they represent robust capabilities and are ideal candidate models for RAG use case deployments.

\begin{table*}[ht]
\centering
\small
\begin{tabular}{llcccccccccc}
\toprule
Model & Method &  \multicolumn{3}{c}{TriviaQA} &  \multicolumn{3}{c}{ASQA} &  \multicolumn{4}{c}{PubmedQA} \\
\cmidrule(lr){3-5} \cmidrule(lr){6-8} \cmidrule(lr){9-12} 
&& EM & Faith. & Rel. & STR-EM & Faith. & Rel. & Acc & F1 & Faith. & Rel. \\
 \midrule
\multirow[l]{5}{*}{Phi-3 3.8B} & Baseline   & 0.630 & - & - & 0.109 & - & - & 0.476& 0.290& - & -\\
& RAG &  0.876 & 0.821 & 0.836 & 0.294 & 0.685 & 0.895 & 0.530 & 0.281 & - & - \\
& RAG-sft & 0.878 & 0.777 & 0.750 & 0.252 &0.717 & 0.833 & \textbf{0.720} & \textbf{0.491} & - & -\\
& CoT & \textbf{0.923} & 0.555 & 0.741 & 0.367 & 0.263 & 0.826 & 0.574 & 0.439 & 0.477 & 0.705 \\
& CoT-sft & 0.795 & 0.793 & 0.749 & \textbf{0.386} & 0.749 & 0.839 & 0.620 & 0.458 & 0.631 & 0.853 \\
\midrule
\multirow[l]{5}{*}{Llama-3 8B} & Baseline  & 0.722 & - & - & 0.200 & - & - & 0.560 & 0.366 & - & -\\
& RAG  & 0.828 & 0.783 & 0.746 & 0.285 & 0.610 & 0.861 & 0.556 & 0.398 & - & - \\
& RAG-sft  & \textbf{0.916} & 0.704 & 0.714 & 0.291 & 0.653 & 0.854 & \textbf{0.770} & \textbf{0.537} & - & -\\
& CoT  & 0.896 & 0.518 & 0.764 & 0.395 & 0.536 & 0.730 & 0.684 & 0.480 & 0.378 & 0.732 \\
& CoT-sft  & 0.851 & 0.808 & 0.697 & \textbf{0.422} & 0.768 & 0.790 & 0.694 & 0.485 & 0.777 & 0.883 \\
\bottomrule
\end{tabular}
\caption{Evaluation results of baseline and different RAG settings, for the three datasets and two models tested. In addition to the main metrics for each dataset, faithfulness and relevancy are reported for the relevant configurations. In bold are the best configurations per dataset, based on the main metrics.}
\label{tab:results}
\end{table*}

\subsection{Evaluation}
We measure and report Exact Match (EM) for \mbox{TriviaQA}, STR-EM for ASQA, accuracy and F1 for PubmedQA. Additionally, we evaluate two \mbox{RAGAS} metrics \cite{es-etal-2024-ragas}: Faithfulness and Relevancy. Faithfulness measures the relation between the generated text and the context. Relevancy measures the relation between the generated text and the query. These two metrics use the context as input for the LLM critic, so are only relevant in the RAG settings. The critic LLM used is GPT4-32k, version 0613. An embedder\footnote{\href{https://huggingface.co/BAAI/bge-small-en-v1.5}{BAAI/bge-small-en-v1.5}.} is required for the relevancy evaluation.

\subsection{Results}

We present a comparative study of RAG augmentation techniques, on the TriviaQA, ASQA and PubmedQA datasets. Results are presented in \cref{tab:results}: main metrics for each dataset are displayed, as well as faithfulness and relevancy scores, as defined in \cite{es-etal-2024-ragas}. For TriviaQA we observe the following: retrieved context improves the results, fine-tuning the RAG setting improves the results, fine-tuning on CoT reasoning (which includes training on a combination of gold passages and distractor passages) decreases performance. Best method is model dependent for this dataset. For ASQA, we similarly observe every method improves upon the baseline, CoT reasoning produces consistent improvement in both models, as well as fine-tuning of the CoT configuration, which shows to perform best. Finally, for PubmedQA, we observe that almost all methods improve upon the baseline (with one exception); CoT reasoning improves upon the untrained RAG setting, but upon fine-tuning, the RAG method appears to perform best in both models. 

Inspecting the faithfulness and relevancy scores, notice that not all configurations are valid to be measured: these metrics require context, so are irrelevant for the baseline method. 
Additionally, in the PubmedQA dataset, the answers are binary Yes/No; only in the CoT configurations the LLMs produce a reasoning, which can be evaluated. Finally, the faithfulness and relevancy scores often do not correlate with the main metrics, neither with each other, possibly indicating they capture different aspects of the retrieval and generated results, and represent a trade-off in performance. 

The results demonstrate the usefulness of RAG techniques for improving performance, as well as the need to carefully evaluate different aspects of a RAG system, on a diverse set of datasets, as effort on developing generalized techniques is ongoing.

\section{Conclusion}

We introduced \ragf, an open-source library dedicated to the task of RAG-augmentation of LLMs, namely fine-tuning LLMs to become better at RAG settings. The library is designed to serve as an end-to-end experimentation environment, enabling users to quickly prototype and experiment with different RAG techniques. We demonstrated the usefulness of the library by augmenting two models with RAG configurations, evaluating on three Q\&A datasets and showing the benefit of RAG techniques, as well as of using multi-aspect metrics relevant for RAG systems evaluation.



\section*{Limitations and Future Plans}
Our hope is that the library will be useful to as many people and use-cases as possible. However, due to time and resource constraint, we were able to demonstrate its usefulness on a subset of tasks and datasets. Future work can expand the evaluation to other tasks, as well as implementing other RAG techniques and evaluations.

Although we designed the library to be general and customizable, there might be specific workflows which will be difficult to run as-is and some code changes may be required. The library proved useful for our own research projects on a diverse set of datasets and tasks and extending it is easy and straightforward. 

Finally, despite our best efforts to offer detailed documentation in the library, there could be some missing details regarding some functionality or specific use-cases. The code repository will accept suggestions, bug-fixes and pull requests.

\section*{Ethics Statement}

In conducting our research we strive abiding to the highest ethical standards, including integrity, fairness, and societal benefit of our work. We prioritized data privacy and security throughout our research; any data used in our experiments was publicly available and did not contain any private information. We are committed to the principles of transparency and reproducibility; the methodologies, including data pre-processing, model training, and evaluation are documented in order to enable others to replicate our findings. Code is made available in an open repository. We advocate for the responsible use of LLMs and RAG augmentation. It is essential to exercise caution and verify the accuracy and reliability of generated text produced by LLMs. Hallucinations can have negative implications, and even when RAG methods can ameliorate some of these aspects, verification and inspections are needed.


\bibliography{anthology,bibliography}

\appendix
\newpage
\clearpage

\section{Implementation Details}

\subsection{Prompts}
\label{appendix:prompts}

\begin{listing}[h!]
\centering
\begin{minted}[fontsize=\small]{text}
You are a helpful question answerer who can provide an answer given a question and relevant context.
\end{minted}
\caption{System instruction used in the experiments.}
\label{fig:system-prompt}
\end{listing}

\begin{listing}[h!]
\centering
\begin{minted}[fontsize=\small]{text}
Question: {query}
Context: {docs}    
\end{minted}
\caption{Template for inserting relevant documents as context.}
\label{fig:prefix-template}
\end{listing}

\begin{listing}[h!]
\centering
\begin{minted}[fontsize=\small]{text}
Question: {query}
Context: {docs}

Answer this question using the information given in the context above. Here is things to pay attention to:
- First provide step-by-step reasoning on how to answer the question.
- In the reasoning, if you need to copy paste some sentences from the context, include them in 
  ##begin_quote## and ##end_quote##. This would mean that things outside of ##begin_quote## and 
  ##end_quote## are not directly copy paste from the context.
- End your response with final answer in the form <ANSWER>: $answer, the answer should be succinct.
\end{minted}
\caption{Template for Chain-of-Thought reasoning.}
\label{fig:cot-template}
\end{listing}

\subsection{Datasets}
\label{appendix:datasets}

Datasets used:
\begin{itemize}[noitemsep,topsep=0.5em,parsep=0.4em,leftmargin=1.5em]
    \item \href{https://huggingface.co/datasets/Tevatron/wikipedia-nq}{TriviaQA}
    \item \href{https://huggingface.co/datasets/din0s/asqa}{ASQA}
    \item \href{https://huggingface.co/datasets/bigbio/pubmed_qa}{PubmedQA}
\end{itemize}
\noindent
Context size was $k=5$, unless indicated otherwise. Dataset sizes are: 

\begin{center}
\begin{tabular}{lcc}
\toprule
Dataset       & Training & Evaluation             \\ 
\midrule
TriviaQA & 6000 & 1000               \\ 
ASQA & 4353 & 948               \\ 
PubmedQA & 10000 & 500 \\
\bottomrule
\end{tabular}
\end{center}

\subsection{Training Details}
\label{appendix:training}

\begin{center}
\begin{tabular}{lcc}
\toprule
Parameter       & Value \\ 
\midrule
LoRA $r$ & 16  \\
LoRA $\alpha$ & 16  \\
LoRA Dropout & 0.1  \\
LoRA Bias & None  \\
LoRA Modules &  \verb|qkv_proj|, Phi-3\\
& \verb|q/v_proj|, Llama-3\\
LR              & 1e-4            \\
LR Scheduler    & \verb|cosine|            \\
Warmup Ratio    & 0.03            \\ 
Weight Decay    & 0.001     \\
Batch Size      & 1               \\
Epochs  & 1            \\ 
\bottomrule
\end{tabular}
\end{center}

\end{document}